\documentclass[10pt,twocolumn,letterpaper]{article}

\usepackage{cvpr}              

\definecolor{cvprblue}{rgb}{0.21,0.49,0.74}
\usepackage[pagebackref,breaklinks,colorlinks,allcolors=cvprblue]{hyperref}
\usepackage{multirow}
\usepackage[accsupp]{axessibility}
\def\paperID{6374} 
\def\confName{CVPR}
\def\confYear{2025}

\usepackage{makecell}
\usepackage[numbers]{natbib}

\usepackage{amsmath}

\usepackage[utf8]{inputenc} 
\usepackage[T1]{fontenc}    
\usepackage{hyperref}       
\usepackage{url}            
\usepackage{booktabs}       
\usepackage{amsfonts}       
\usepackage{nicefrac}       
\usepackage{microtype}      
\usepackage{xcolor}         

\usepackage{url}
\usepackage{times}
\usepackage{epsfig}
\usepackage{amsmath}
\usepackage{amssymb}
\usepackage{graphicx}
\usepackage{multirow}
\usepackage{multicol}
\usepackage{algorithmic}
\usepackage{booktabs}
\usepackage{enumitem}
\usepackage{colortbl}  
\usepackage{array}   
\usepackage{pifont}
\usepackage{xspace}

\title{UCOD-DPL: Unsupervised Camouflaged Object Detection \\ via Dynamic Pseudo-label Learning}
\author{
\normalsize Weiqi Yan$^{1}$, Lvhai Chen$^{1}$, Huaijia Ko$^{1}$, Shengchuan Zhang$^{1}$\thanks{Corresponding author}, Yan Zhang$^{1}$, Liujuan Cao$^{1}$\\
\normalsize $^{1}$ Key Laboratory of Multimedia Trusted Perception and Efficient Computing, \\
\normalsize  Ministry of Education of China, Xiamen University, 361005, P.R. China. \\
{\tt\small weiqi\_yan@outlook.com\ lvhaichen2002@gmail.com\ zsc\_2016@xmu.edu.cn}
}

\begin{document}
\maketitle
\vspace{-1cm}
\begin{abstract}
Unsupervised Camoflaged Object Detection (UCOD) has gained attention since it doesn't need to rely on extensive pixel-level labels. Existing UCOD methods typically generate pseudo-labels using fixed strategies and train $1 \times 1$ convolutional layers as a simple decoder, leading to low performance compared to fully-supervised methods. We emphasize two drawbacks in these approaches: 1). The model is prone to fitting incorrect knowledge due to the pseudo-label containing substantial noise. 2). The simple decoder fails to capture and learn the semantic features of camouflaged objects, especially for small-sized objects, due to the low-resolution pseudo-labels and severe confusion between foreground and background pixels. To this end, we propose a UCOD method with a teacher-student framework via Dynamic Pseudo-label Learning called UCOD-DPL, which contains an Adaptive Pseudo-label Module (APM), a Dual-Branch Adversarial (DBA) decoder, and a Look-Twice mechanism. The APM module adaptively combines pseudo-labels generated by fixed strategies and the teacher model to prevent the model from overfitting incorrect knowledge while preserving the ability for self-correction; the DBA decoder takes adversarial learning of different segmentation objectives, guides the model to overcome the foreground-background confusion of camouflaged objects, and the Look-Twice mechanism mimics the human tendency to zoom in on camouflaged objects and performs secondary refinement on small-sized objects. Extensive experiments show that our method demonstrates outstanding performance, even surpassing some existing fully supervised methods.
 The code is available now\footnote{\url{https://github.com/Heartfirey/UCOD-DPL}}. 
\end{abstract}

\section{Introduction}
\label{sec:intro}
``Camouflage'' originates from the natural behavior of objects blending into their surroundings using similar textures and colors to evade ``predators''\cite{caro2005adaptive}. Camouflaged Object Detection (COD) is a challenging semantic segmentation task aimed at learning to segment objects that are visually hidden in their backgrounds, which has significant applications in several important fields \cite{fan2020pranet,tabernik2020segmentation}.

\begin{figure}
    \centering
    \includegraphics[width=\linewidth]{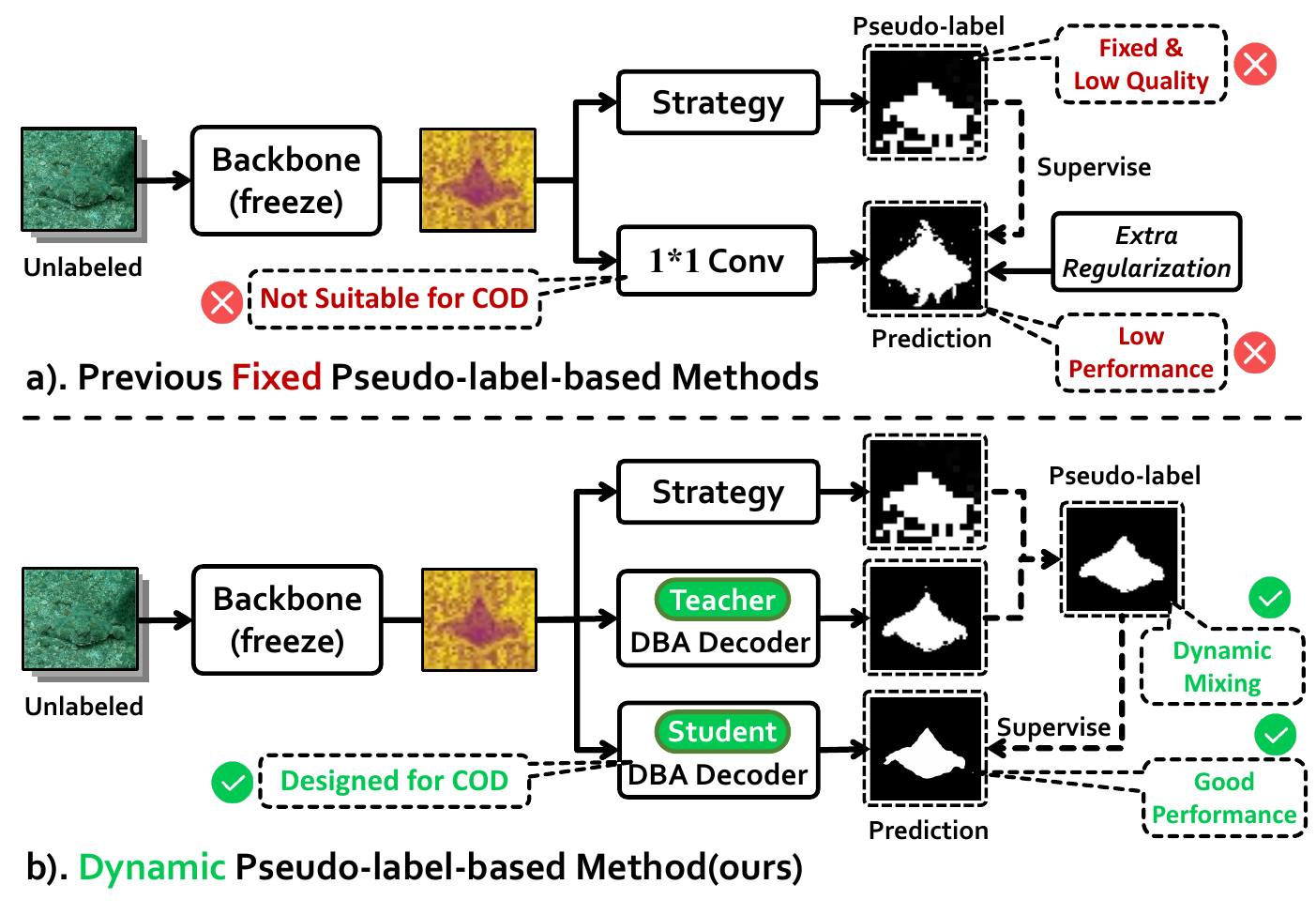}
    \caption{Comparison between our method and previous pseudo-label-based methods.}
    \label{fig:cover}
\end{figure}

Although fully-supervised methods have achieved significant progress in the COD tasks \cite{yin2024camoformer,pang2023ZoomNextPAMI, yao2024hierarchical,sun2024glconet,he2023camouflaged}, they heavily rely on large-scale human annotations for training. However, it is difficult to get pixel-level human labels for camouflaged objects due to their complex visual attributes \cite{lai2024camoteacher}. Therefore, this reliance on annotations has sparked increasing interest in the task of unsupervised camouflage object detection (UCOD) among researchers. However, existing UCOD methods typically use pseudo-labels generated by fixed strategies to train a single $1 \times 1$ convolutional layer \cite{simeoni2023unsupervised, zhang2023unsupervisedco}. By investing in existing methods, we identified two main issues: 1). The pseudo-label constructed by fixed strategies contains substantial noise, which makes the model prone to fitting incorrect knowledge. 2). These pseudo-labels have low resolution and severe confusion between foreground and background pixels, and simple $1 \times 1$ convolution fails to capture and learn the semantic features of camouflaged objects, especially for small-sized objects. Consequently, their segmentation results fall far behind those of fully supervised methods, which limits their application in real-world scenarios. 

\begin{figure}
    \centering
    \includegraphics[width=\linewidth]{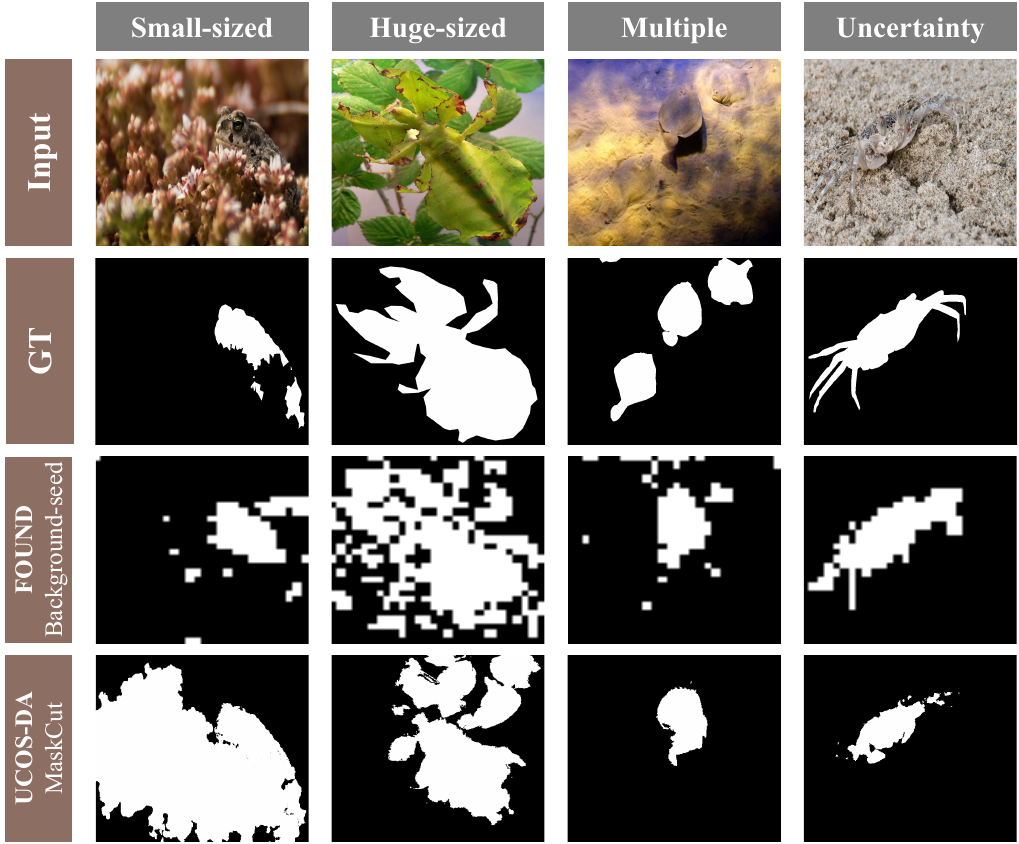}
    \caption{\textbf{Some examples of low-quality pseudo-labels generated using fixed strategies.} Based on features extracted by \textit{DINOv2}, we construct pseudo-labels using the Background-Seed from FOUND \cite{simeoni2023unsupervised} and the MaskCut in UCOD-DA \cite{simeoni2023unsupervised}, in challenging scenarios (\textit{e.g.} small-sized objects, huge-sized objects, multiple instances, and uncertain objects) for visualization.}
    \label{fig:cover_pseudo_label}
\end{figure}

In this paper, we propose a brand-new model called UCOD-DPL. As shown in \cref{fig:cover}, we introduce a teacher-student framework consisting of an Adaptive Pseudo-label Mixing (APM) module, a Dual-Branch Adversarial (DBA) decoder, and a Look-Twice refinement strategy. As illustrated in \cref{fig:cover_pseudo_label}, pseudo-labels constructed by fixed strategies are generally low-quality and contain substantial noise. To prevent the model from fitting uncorrectable erroneous knowledge in the pseudo-labels, we design the APM module with a discriminator and a dynamic scoring mechanism to mix the pseudo-labels from the fixed strategy and the teacher branch adaptively. To help the model overcome foreground-background confusion, we propose a DBA decoder inspired by predator-prey dynamics. This decoder uses different segmentation targets (foreground and background) along with an orthogonal loss of features to deeply mine and learn distinctive features of the foreground and background. Additionally, inspired by \cite{pang2022ZoomNetCVPR2022, pang2023ZoomNextPAMI}, we mimic the zooming-in behavior humans use when examining camouflaged images by performing secondary refinement on small-sized camouflaged objects to enhance the model's segmentation accuracy when dealing with small targets.

In the experiment, we train our model on the commonly used COD training set and evaluate it on multiple test sets. The results demonstrate that our method achieves outstanding performance, even comparable to some fully supervised COD methods when using \textit{DINOv2} as the backbone.

Our main contributions are summarized as follows:
\begin{enumerate}
    \item We propose a novel UCOD model called UCOD-DPL, and extensive experiments demonstrate that our method exhibits superior performance, reaching the performance level of some fully supervised algorithms.
    \item We introduce a teacher-student framework with an APM module. By using the APM, we ensure that the model can learn knowledge from pseudo-labels while preserving the ability to identify and correct the erroneous.
    \item To address the issues of low-resolution pseudo-labels and foreground-background confusion, we design a DBA decoder that uses a foreground-background adversarial task, enabling the model to learn the camouflaged features of the foreground adaptively.
    \item To enhance the model’s ability to segment small-sized objects, we design a Look-Twice refinement mechanism. This mechanism mimics human behavior by performing a secondary segmentation on smaller camouflaged objects, thereby refining the segmentation results.
\end{enumerate}

\section{Related Works}
\label{sec:formatting}

\subsection{Unsupervised Semantic Segmentation}

Unsupervised Semantic Segmentation (USS) aims to train a model without any human annotations to produce the semantic segmentation mask. Before the surge of deep learning methods, researchers relied on numerous handcrafted methods \cite{goferman2011context,yan2013hierarchical,zhu2014saliency,cheng2014global} that used one or more priors related to foreground regions in the image to generate segmentation masks, such as center\cite{judd2009learning}, contrast\cite{itti1998model}, and boundary\cite{wei2012geodesic}. In deep learning-based methods, some approaches utilize generative models \cite{goodfellow2014generative} to produce a segmentation mask, which will be used to synthesize a realistic image by copying the corresponding region into a background which is either synthesized \cite{bielski2019emergence} or taken from a different image \cite{arandjelovic2019object}. Another category of methods leverages pseudo-labels constructed by handcrafted fixed strategies to train the detector. \cite{zhang2018deepusd} adopts diverse noisy pseudo-labels via distinct unsupervised handcrafted methods to train the detector to predict a saliency map free from noise in labels. And \cite{nguyen2019deepusps, simeoni2023unsupervised} proposed to refine the pseudo-labels produced by fixed-strategy by training the model via a self-supervised iterative refinement manner. These methods adopt a self-supervision manner that highly relies on pseudo-labels generated directly from one or multiple fixed strategies. If the pseudo-labels contain substantial noise, these methods remains prone to fitting incorrect knowledge, which is particularly evident when dealing with camouflaged objects.

\subsection{Camouflaged Object Detection}
Camouflaged Object Detection (COD) is a complex task aimed at segmenting objects that are intentionally designed to merge seamlessly with their environment. This task has an extensive history in its domain. Existing fully-supervised COD methods \cite{pang2022ZoomNetCVPR2022, pang2023ZoomNextPAMI, fan2020camouflaged, ZHUGE2022108644} have achieved excellent performance by training on large-scale annotated datasets. These methods utilize the rich information (\textit{e.g.} boundary \cite{ji2023gradient, Zhu2022ICF, sun2023edgeaware}, texture \cite{ji2023gradient, Zhu2021InferringCO, ren2023dtfcodieee} and other information like frequency domain and depth \cite{zhong2023cvprdcoifd, Lin2023FACOD-, Wang2023DACOD, Xiang2021ExploringDC}) provided by pixel-level annotations and design a series of sophisticated strategies to extract unique visual characteristics of camouflaged objects. However, the extensive pixel-level annotations are hardly gained, which has drawn the attention of researchers to Unsupervised Camouflaged Object Detection (UCOD) and USS methods \cite{zhang2023unsupervisedco}. These methods typically rely on pseudo-labels generated through fixed strategies to supervise model learning for segmentation. However, due to limitations in algorithmic complexity and robustness, these fixed strategies struggle to work effectively at high resolutions and easily introduce significant noise. Consequently, these methods are prone to learning incorrect noise from the pseudo-labels, leading to suboptimal segmentation performance and limiting the application in the real world of UCOD models.

\subsection{Self-supervised Learning}
Self-Supervised Learning (SSL) has revolutionized the way models leverage unlabeled data. Among current SSL approaches, some methods adopt contrastive learning to learn image feature representations. MoCo \cite{he2020momentum} utilizes a dynamic memory bank to maintain a large pool of negative samples; SimCLR \cite{chen2020simple} employs various data augmentations and a projection head to maximize similarities between differently augmented views of the same instance; and BYOL \cite{grill2020bootstrap} constructs a teacher-student framework with the teacher model's parameters are updated by Exponential Moving Average (EMA) of the student model. Following BYOL, DINO \cite{caron2021emerging} applied two interactive encoders sharing the same structure but with different parameter sets and update strategies. Inspired by self-supervised pretraining techniques in natural language processing \cite{sun2019bert4rec,radford2018improving,radford2019language}, other SSL methods design pretext tasks (\textit{e.g.} colorization, rotation prediction, and patch reconstruction) on unlabeled images \cite{caron2021emerging,chen2020simple,gidaris2021obow,gidaris2018unsupervised,he2020momentum,noroozi2016unsupervised,lin2024weakly,mi2022active}. BeiT \cite{bao2021beit} adopts a Transformer-based architecture to reconstruct masked image patches, MAE \cite{he2022masked} implements a masked autoencoder that efficiently reconstructs the missing parts of an image. In this paper, we employ \textit{DINOv1} \cite{caron2021emerging} and \textit{DINOv2} \cite{oquab2023dinov2} to extract robust and generalized image features.

\section{Methodology}
\label{sec:methods}

\begin{figure*}[!t]
    \centering
    \includegraphics[width=\linewidth]{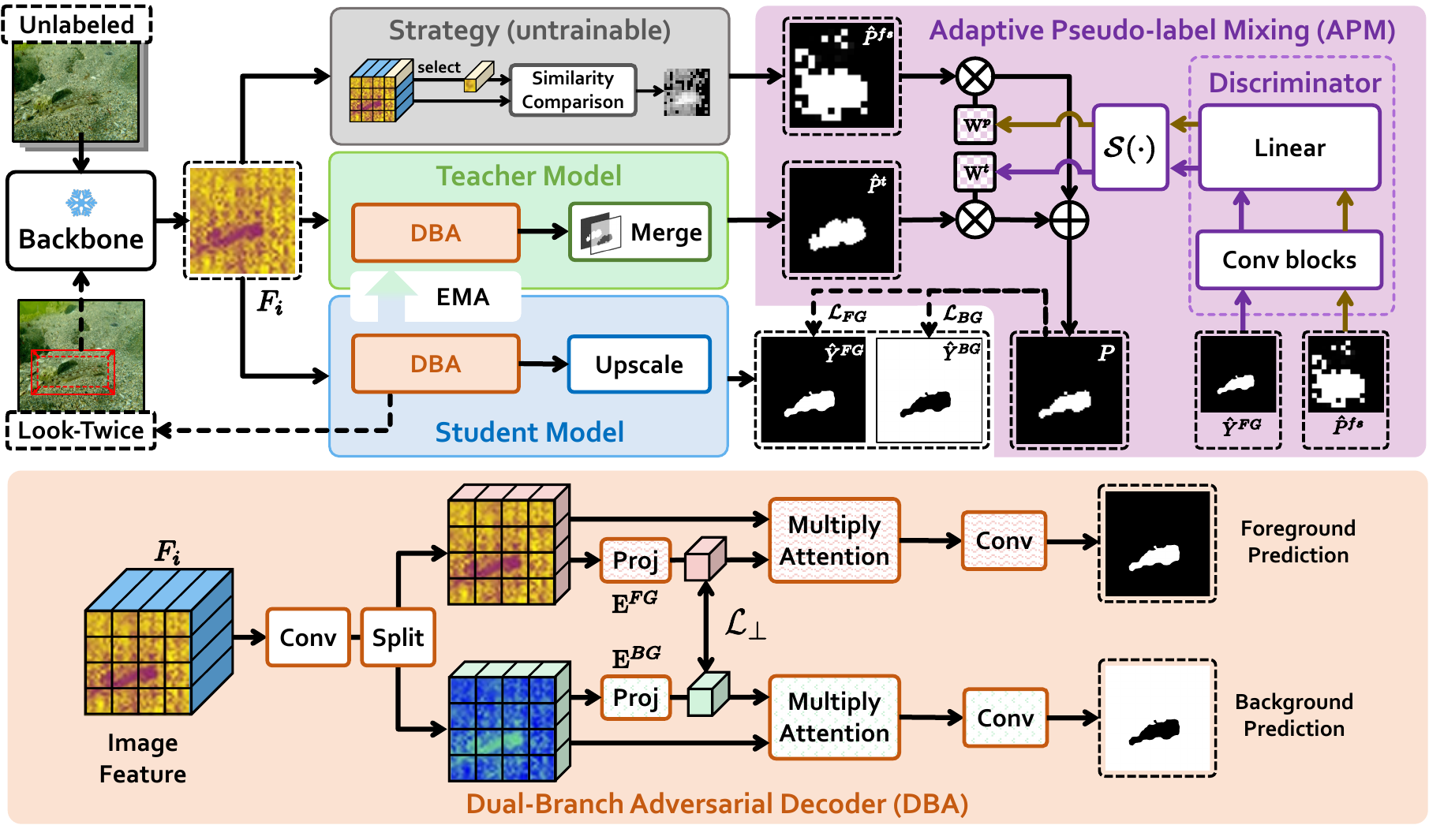}
    \caption{\textbf{The main framework of our proposed method.} The model contains a teacher-student framework, an Adaptive Psdudo-label Mixing (APM) module, a Dual-Branch Adversarial (DBA) decoder, and a Look-Twice strategy.}
    \label{fig:framework}
\end{figure*}

\subsection{Overall Framework}
The structure of our proposed method is shown in \cref{fig:framework}, which includes a teacher-student framework with an Adaptive Pseudo-label Mixing (APM) module, a Dual-Branch Adversarial (DBA) decoder, and a Look-Twice strategy. Both the teacher and student models are configured as DBA decoders.

Firstly, the $i$-th unlabeled input image passes through the self-supervised pretrained backbone to obtain high-level semantic image feature ${F}_i$. We then employ the background seed method to generate a low-resolution fixed-strategy pseudo-label $\hat{P}^{fs}_i$, which utilizes pixel-level similarity comparison and was proposed in \cite{simeoni2023unsupervised}. Simultaneously, the image feature $F_i$ is fed into a teacher-student model. The teacher model generates a higher-resolution teacher's pseudo label $\hat{P}^{t}_i$, and the student model produces the foreground and background predicted mask $\hat{Y}^{FG}_i, \hat{Y}^{BG}_i$. Then, $\hat{P}^{fs}_i$ and $\hat{P}^t_i$ are fed into the APM module to construct the mixed dynamic pseudo-label $P_i$. Finally, we use the mixed dynamic pseudo-label $P_i$ to supervise the final foreground and background predicted masks $\hat{Y}^{FG}_i, \hat{Y}^{BG}_i$. We further apply a Look-Twice strategy to perform a secondary refinement on small-sized objects to improve the model's segmentation performance on small-sized samples.

\subsection{Adaptive Pseudo-Label Merging Module}
\label{sec:apm}
Existing unsupervised semantic segmentation methods typically adopt a fixed-strategy (\textit{e.g.} similarity comparison \cite{shin2022unsupervised,wang2022selftfu,melas2022deep}) or utilize a dual-branch model (\textit{e.g.} fixed-strategy or learnable branch \cite{zhang2023unsupervisedco,simeoni2023unsupervised}) structure for self-supervised learning to generate segmentation masks. However, we believe the performance of these methods is constrained by the strategy's design or the pseudo-labels' quality, preventing the model from achieving higher accuracy. 

We introduce a discriminator $\mathcal{D}$, which inputs a foreground segmentation mask (\textit{i.e.} fixed-strategy pseudo-label $\hat{P}^{fs}_i$ or student's foreground predicted mask $\hat{Y}^{FG}_i$) and outputs the probability that the mask originates from the fixed-strategy branch:
\begin{equation}
\begin{aligned}
    \mathrm{\hat{y}}^{p1}_i = \mathcal{D}(\hat{P}^{fs}_i),\ \mathrm{\hat{y}}^{p2}_i = \mathcal{D}(\hat{Y}^{FG}_i),
\end{aligned}
\end{equation}
where $\hat{\mathrm{y}}^{p1}_i, \hat{\mathrm{y}}^{p2}_i$ denote the predicted probability that the input mask belongs to the fixed-strategy branch. We then introduce a scoring function $\mathcal{S}(\cdot)$ with a temporal constraint: \vspace{-0.2cm}
\begin{equation}
\begin{aligned}
    \mathcal{S}(\hat{\mathrm{y}}^{p1}_i, \hat{\mathrm{y}}^{p2}_i) = \mathrm{CLIP} \left(\frac{t}{T} + \frac{1}{2}(1 + \cos(\pi \times |\mathrm{\hat{y}}^{p1}_i - \mathrm{\hat{y}}^{p2}_i|)\right)_0^1,
\end{aligned}
\end{equation}
where $t, T$ denote the current and total training epochs, $\mathrm{CLIP}(x)_0^1$ denotes truncating $x$ so that it lies within the interval $[0,1]$. In the early stage of training, the teacher model has not yet developed sufficient localization and basic semantic segmentation capabilities for camouflaged objects, making it unable to supervise the student model effectively. At this point, there is a significant difference between fixed-strategy pseudo-label $\hat{P}^{fs}_i$ and the student's foreground predicted mask $\hat{Y}^{FG}_i$. To help the model learn the fundamental foreground localization and segmentation tasks, we use a higher proportion of $\hat{P}^{fs}_i$ to supervise model training. Later in the training process, as the $\hat{Y}^{FG}_i$ gradually approaches $\hat{P}^{fs}_i$, we increase the proportion of the teacher's pseudo-label $\hat{P}^{t}_i$ as it can make more stable predictions. This adjustment helps prevent the model from overfitting to noise and incorrect knowledge in the $\hat{P}^{fs}_i$. We use the obtained score as the mixing weight $\mathrm{W}^{t}_i$ to combine the fixed-strategy pseudo-label $\hat{P}^{fs}_i$ and teacher's predictions $\hat{P}^{t}_i$:
\begin{equation}
\begin{aligned}
    &\mathrm{W}^{t}_i = \mathcal{S}(\hat{\mathrm{y}}^{p1}_i, \hat{\mathrm{y}}^{p2}_i), \\
    &P_i = \mathrm{W}^{t}_i \hat{P}^{t}_i + (1 - \mathrm{W}^{t}_i) \hat{P}^{fs}_i,
\end{aligned}
\end{equation}
where $P_i$ denotes the mixed dynamic pseudo-label. For the training of the discriminator, we set a binary classification task by using $\hat{P}^{fs}_i$ and $\hat{Y}^{FG}_i$ as the input and assign the discriminator label $\mathrm{y}^{p1}_i = 1$ for the fixed-strategy pseudo-label $\hat{P}^{fs}_i$ and $\mathrm{y}^{p2}_i=0$ for the student model branch $\hat{Y}^{FG}_i$. Then, we employ a cross-entropy supervision loss $\mathcal{L}_{\mathrm{BCE}}$ to train the discriminator:
\begin{equation}
    \mathcal{L}_{dis} = \mathcal{L}_{\mathrm{BCE}}(\hat{\mathrm{y}}^{pj}_i, \mathrm{y}^{pj}_i), j \in \{1, 2\}.
    \label{eq:loss_dis}
\end{equation}

During the training, we treat the student model as a kind of generator and alternately train the decoder and the discriminator in a GAN-like manner.

\subsection{Dual-Branch Adversarial Decoder}
In previous methods \cite{simeoni2023unsupervised,  zhang2023unsupervisedco}, predictions are typically generated by a simple $1 \times 1$ convolution, which limits the ability to capture camouflage features thoroughly. We propose a Dual-Branch Adversarial decoder to enhance accuracy and robustness in this task. The DBA decoder consists of two parallel branches: one is responsible for segmenting the foreground predicted mask $\hat{Y}^{FG}_i$, while the other is for segmenting the background predicted mask $\hat{Y}^{BG}_i$.

First, for $i$-th image feature ${F}_i$ extracted from the backbone model, a convolutional layer is used to double its channel, then we decouple it into foreground and background features ${F}_i^{FG}, {F}_i^{BG}$ with the same shape of ${F}_i$:
\begin{equation}
    {F}_i^{FG}, {F}_i^{BG} = \mathrm{Split}(\mathrm{Conv}({F}_i)),
\end{equation}
where $\mathrm{Split}(\cdot)$ denotes splitting a tensor along the channel dimension equally. We then employ two learnable embeddings $\mathrm{E}_{FG}, \mathrm{E}_{BG}$ to store foreground and background-related knowledge. These embeddings are used to calculate foreground and background attention queries ${Q}_i^{FG'}, {Q}_i^{BG'}$:
\begin{equation}
    {Q}_i^{FG'} = {F}_i^{FG} \times \mathrm{E}_{FG},\ {Q}_i^{BG'} = {F}_i^{BG} \times \mathrm{E}_{BG}.
\end{equation}

Finally, we apply multiplicative attention with a residual connection to weigh the features, enhancing important details. The resulting features are then passed through a $1 \times 1$ convolution to output the foreground predicted mask $\hat{Y}_i^{FG}$ and background mask $\hat{Y}_i^{BG}$:
\begin{equation}
\begin{aligned}
    \hat{Y}_i^{FG} = \mathrm{Conv'}(\mathrm{Sigmoid}({Q}_i^{FG'} \times {F}_i^{FG}) + {F}_i^{FG}), \\
    \hat{Y}_i^{BG} = \mathrm{Conv'}(\mathrm{Sigmoid}({Q}_i^{BG'} \times {F}_i^{BG}) + {F}_i^{BG}). \\
\end{aligned}
\end{equation}

For the student model, we use $Y_i^{FG}, \hat{Y}_i^{BG}$ as the student's foreground and background prediction masks. For the teacher model, we further generate the teacher's pseudo-label $P_i$ using the combination of the foreground and background prediction masks.

Inspired by \cite{wu2021vector}, we apply an orthogonal loss $\mathcal{L}_{\bot}$ for ${Q}_i^{FG'} \in \mathbb{R}^{(n\times m) \times c}, {Q}_i^{BG'} \in \mathbb{R}^{(n \times m) \times c}$ as a regularisation constraint to encourage them to focus on distinct features:
\begin{equation}
\begin{aligned}
    \mathbb{M} &= {Q}_i^{FG'} ({Q}_i^{BG'})^{\top}, \\
    \mathcal{L}_{\bot} = \frac{1}{(n \times m)^2} &\sum_{j = 1}^{n\times m} \sum_{k=1}^{n \times m}(1 - \delta_{jk})(\mathbb{M}[j, k])^2,
    \label{eq:loss_org}
\end{aligned}
\end{equation}
where $n, m, c$ indicate the height, width, and number of channels. When $j = k$, $\delta_{jk} = 1$; otherwise $\delta_{jk} = 0$. 
For the predictions $\hat{Y}_i^{FG}, \hat{Y}_i^{BG}$, we use a binary cross-entropy loss $\mathcal{L}_{\mathrm{BCE}}$ for supervision, where the background prediction is inverted before applying the supervision:
\begin{equation}
    \mathcal{L}_{seg} = \mathcal{L}_{\mathrm{BCE}}(\hat{Y}^{FG}_i, P_i) + \mathcal{L}_{\mathrm{BCE}}\left((1 - \hat{Y}^{BG}_i), P_i\right).
    \label{eq:loss_seg}
\end{equation}

\subsection{Look-Twice Refinement}
\label{subsec:look_twice}

The segmentation quality of COD is often influenced by object size. Due to reduced spatial resolution, the model struggles to produce fine segmentation for small-sized camouflaged objects. Inspired by previous works \cite{pang2022ZoomNetCVPR2022, pang2023ZoomNextPAMI}, we attempt to mimic human behavior when observing small-sized objects: first coarsely locating the camouflaged object in the image, then zooming in to carefully retrieve its details. To this end, we designed the Look-Twice mechanism. First, the input image is processed by the model to obtain a coarse segmentation result $\hat{Y}^{FG}_i$. We define the set of all foreground objects as $\mathrm{C}^{FG}$, which can be calculated as:
\begin{equation}
    \mathrm{C^{FG}} = \mathbf{ConnectComponent}(\hat{Y}^{FG}_i),
\end{equation}
where $\mathbf{ConnectComponent}(\cdot)$ denotes the connected components labelling algorithm. Next, for each connect component $c_k^{FG} \in \mathrm{C}^{FG}$, we calculate its foreground area ratio $r_k^{FG}$:
\begin{equation}
    \begin{aligned}
        r_k^{FG} = \frac{\sum_{(x, y) \in \mathrm{c}_k^{FG}} \hat{Y}^{FG}_i(x, y)}{H \times W},
    \end{aligned}
\end{equation}
where $H, W$ denote the height and width of the foreground area of the output mask $\hat{Y}^{FG}_i$.
We consider all regions where $r_i^{FG} < \tau$ as small-sized object areas. We set $\tau = 0.15$ when using \textit{DINOv2} as the backbone. The hyper-parameter ablation study on $\tau$ will be provided in the experiments section. 

For each cropped small-sized object region, we aim to retain a certain proportion of the background to provide the model with sufficient context to segment the foreground from the background. Therefore, for each connected component region $\mathrm{c}_k^{FG}$, we calculate the local foreground area ratio $s_k^{FG}$ and the global area ratio $s_k^{BG}$:
\begin{equation}
\begin{aligned}
    &s_k^{FG} = \frac{\sum_{(x, y) \in \mathrm{c}_k^{FG}} \hat{Y}^{FG}_i(x, y)}{H_k^{FG} \times W_k^{FG}},  \\
    &s_k^{BG} = \frac{H_k^{FG} \times W_k^{FG}}{H \times W},
\end{aligned}
\end{equation}
where $H_k^{FG}, W_k^{FG}$ denote the height and width of the bounding box of the $k$-th connected component $\mathrm{c}_k^{FG}$:
\begin{equation}
    \mathrm{scale}_k^{FG} = 1 - ({s_k^{FG}}/{s_k^{BG}}).
    \label{eq:lt_scale}
\end{equation}

Finally, we calculate the expansion ratio using \cref{eq:lt_scale} and enlarge the bounding box of the connected component region by this ratio. The expanded bounding box is then used to crop the related area of the input image, and the cropped patches are resized to the original input size. During the training process, these patches are treated as augmented data. At test time, we re-infer these patches, scale the resulting foreground predicted mask back to the original size, and paste it into the coarse mask as the refined result. However, we found that this mechanism still has some weakness: for cases where objects are occluded and the annotations are fragmented, this method may mistakenly amplify a part of the object as a small object. This can serve as a point of our future work.

\subsection{Total Loss}
\label{subsec:total_loss}
Finally, the total loss $\mathcal{L}_{tot}$ can be defined as follows:
\begin{equation}
    \mathcal{L}_{tot} = \mathcal{L}_{seg} + \mathcal{L}_{\bot} + \mathcal{L}_{dis},
\end{equation}
where $\mathcal{L}_{seg}$ is the segmentation loss defined by \cref{eq:loss_seg}, $\mathcal{L}_{\bot}$ is the orthogonal loss defined by \cref{eq:loss_org}, and $\mathcal{L}_{dis}$ is the discriminator loss defined by \cref{eq:loss_dis}. When training on the teacher-student framework, we freeze the parameters of the discriminator and use $\mathcal{L}_{seg}$ to supervise the model training. Conversely, when training the discriminator, we freeze the parameters of the other parts of the model and train the discriminator only using $\mathcal{L}_{dis}$.

\def\metricsCOD{
    &$\mathcal{S}_m\uparrow$
    &$\mathcal{F}_{\beta}^{\omega}\uparrow$
    &$\mathcal{F}_{\beta}^{m}\uparrow$
    &$\mathcal{E}_{\phi}^{m}\uparrow$
    
    &$\mathcal{M}\downarrow$
}
\definecolor{myGray}{gray}{.92}
\begin{table*}[t!]
    \setlength{\belowcaptionskip}{0cm}   
    \renewcommand{\arraystretch}{1.0}
    \renewcommand{\tabcolsep}{2pt}
    \footnotesize
    \centering
\resizebox{\linewidth}{!}{
	\begin{tabular}{c|ccccc|ccccc|ccccc|ccccc}
        \toprule
        \multicolumn{1}{c}{\multirow{2}{*}[-1.2ex]{\textbf{\large Methods}}} & \multicolumn{5}{c}{\textbf{CHAMELEON (87)}} & \multicolumn{5}{c}{\textbf{CAMO-Test (250)}}  & \multicolumn{5}{c}{\textbf{COD10K-Test (2,026)}}& \multicolumn{5}{c}{\textbf{NC4K (4,121)}}\\
        \cmidrule[0.05em](lr){2-6} \cmidrule[0.05em](lr){7-11} \cmidrule[0.05em](lr){12-16} \cmidrule[0.05em](lr){17-21}
        \metricsCOD{}
        \metricsCOD{}
        \metricsCOD{}
        \metricsCOD{} \\
        \midrule
        \multicolumn{21}{c}{\textit{\textbf{Fully-Supervised Methods}}} \\ \midrule
		SINet$_{20}$\cite{fan2020camouflaged} &.872&.806&.827&.946& .034 & .751 & .606 & .675 & .771  & .100 & .771 & .551 & .634 & .806  & .051 & .808 & .723 & .769 & .871  & .058 \\
        $\mathrm{C^2}$FNet$_{21}$\cite{sun2021context} &.888&.828&.844&.946& .032 & .796 & .719 & .762 & .864  & .080 & .813& .686 & .723 & .900  & .036 & .838 & .762 & .794 & .904  & .049 \\
        MGL-R$_{21}$\cite{Mzhai2021mutual} &.893&.812&.834&.941&.030  & .775 & .673 & .726 & .842  & .088 & .814& .666 & .710 & .890  & .035 & .833 & .739 & .782 & .893  & .053 \\
        UGTR$_{21}$\cite{yang2021uncertainty} &.887&.794&.819&.940&.031  & .784 & .684 & .735 & .851  & .086 & .817& .666 & .711 & .890  & .036 & .839 & .746 & .787 & .899  & .052 \\       
        BGNet$_{22}$\cite{sun2022boundary} &.901&.851&.860&.954& .027 & .812 & .749 & .789 & .870  & .073 & .831 & .722 & .753 & .901 & .033 & .851 & .788 & .820  & .907 & .044 \\
        ZoomNet$_{22}$\cite{pang2022ZoomNetCVPR2022} &.902&.845&.864&.958&.023 & .820 & .752 & .794 & .878 & .066 & .838 & .729 & .766 & .888  & .029 & .853 & .784 & .818  & .896 & .043 \\
        SINetv2$_{22}$\cite{fan2021concealed} &.888&.816&.835&.961&.030 & .820 & .743 & .782 & .882   & .070 & .815 & .680 & .718 & .887  & .037 & .847 & .770 & .805  & .903 & .048 \\
        HitNet$_{23}$\cite{hu2023high} &\underline{.921}&\underline{.897}&\underline{.900}&.\textbf{972}&\underline{ .019}& .849 & \underline{.809} &\underline{ .831} &\underline{ .906}  & .055 & \underline{.871} & \underline{.806} & \underline{.823} & \underline{.935}  & \underline{.023} & \underline.875 & \underline{.834} & \underline{.853} & \underline{.926} & .037 \\
        FSPNet$_{23}$\cite{huang2023feature} &.908&.851&.867&.965&.023 & \underline{.856} & .799 & .830 & .899   & \underline{.050} & .851 & .735 & .769 & .895  & .026 & \underline{.879} & .816 & .843  & .915 & \underline{.035} \\ 
        BiRefNet$_{24}$\cite{zheng2024bilateral} &\textbf{.929}&\textbf{.911}&\textbf{.922}&\underline{.968}&\textbf{.016}  & \textbf{.932} & \textbf{.914} &  \textbf{.922} & \textbf{.974} & \textbf{.015} & \textbf{.913} & \textbf{.874} & \textbf{.888}  &  \textbf{.960}   & \textbf{.014} & \textbf{.914}   &    \textbf{.894}   &  \textbf{.909}  &  \textbf{.953}   & \textbf{.023} \\ \midrule
        \multicolumn{21}{c}{\textit{\textbf{Semi-Supervised Methods}}} \\ \midrule
        CamoTeacher$_{24}$(1\%)\cite{lai2024camoteacher} & .652 & .472 & .558 & .714  & .093  & .621 & .456 & .545 & .669  & .136 & .699 & .517 & .788 & .797  & .062 & .718 & .599 & .779 & .814 & .090\\
        CamoTeacher$_{24}$(5\%)\cite{lai2024camoteacher} & .729 & .587 & .656 & .785  & .070 & .669 & .523 & .601 & .711  & .122 & .745 & .583 & .827 & .840  & .050 & .777 & .677 & .834  & .859  & .071\\
        CamoTeacher$_{24}$(10\%)\cite{lai2024camoteacher} & .756 & .617 & .684 & .813  & .065 & .701 & .560 & .742 & .795 & .112 & .759 & .594 & .836 & .854  & .049 & .791 & .687 & .842 & .868  & .068\\
        SCOD-ND$_{24}$(10\%)\cite{fu2024semisupervised} & .850 & .773 & - & .928  & .036 & .789 & .732 & - & .859  & .077 & .819 & .725 & - & .891  & .033 & .838 & .787 & - & .903  & .046\\
        \midrule
        \multicolumn{21}{c}{\textit{\textbf{Unupervised Methods}}} \\ \midrule
        BigGW$_{21}$ \cite{voynov2021oswl}& .547 & .244 & .294 & .527  & .257 & .565 & .299 & .349 & .528  & .282 & .528 & .185 & .246 & .497  & .261 & .608 & .319 & .391 & .565  & .246 \\
        TokenCut$_{22}$\cite{wang2022selftfu} & .654 & .496 & .536 & .740  & .132 & .633 & .498 & .543 & .706  & .163 & .658 & .469 & .502 & .735  & .103 & .725 & .615 & .649 & .802  & .101\\
        TokenCut$_{22}$ w/B.S.\cite{wang2022selftfu} & .655 & .351 & .393 & .582  & .169 & .639 & .383 & .434 & .595  & .195 & .666 & .334 & .399 & .609  & .127 & .735 & .478 & .547 & .683  & .133\\
        SpectralSeg$_{22}$\cite{melas2022deep} & .575 & .410 & .440 & .628  & .220 & .579 & .450 & .481 & .648  & .235 & .575 & .360 & .388 & .595  & .193 & .669 & .535 & .562 & .719  & .159 \\
        SelfMask$_{22}$\cite{shin2022unsupervised} & .619 & .436 & .481 & .675  & .176 & .617 & .483 & .536 & .698  & .176 & .637 & .431 & .469 & .679  & .131 & .716 & .593 & .634 & .777  & .114\\
        SelfMask$_{22}$ w/U.B.\cite{shin2022unsupervised} & .629 & .447 & .491 & .683  & .169 & .627 & .495 & .547 & .708  & .182 & .645 & .440 & .478 & .687  & .125 & .723 & .601 & .642 & .784  & .110  \\
        FOUND$_{23-\textit{DINOv1}}$\cite{simeoni2023unsupervised} & .684 & .542 & .590 & .810  & .095 & .685 & .584 & .633 & .782  & .129 & .670 & .482 & .520 & .751  & .085 & .741 & .637 & .674 & .824  & .084\\
        \textbf{*}FOUND$_{\textit{23-DINOv2}}$\cite{simeoni2023unsupervised}&\underline{.829}&\underline{.757}&\underline{.781}&\underline{.911}&\underline{.040}&.\underline{770}&\underline{.704}&\underline{.740}&\underline{.849}&\underline{.090}&\underline{.767}&\underline{.641}&\underline{.668}&\underline{.847}&\underline{.045}&\underline{.816}&\underline{.756}&\underline{.783}&\underline{.893}&\underline{.052}\\
        UCOS-DA$_{23-\textit{DINOv1}}$\cite{zhang2023unsupervisedco} & .715 & .591 & .629 & .802 & .095 & .701 & .606 & .646 & .784  & .127 & .689 & .513 & .546 & .740  & .086 & .755 & .656 & .689 & .819  & .085\\
        \textbf{*}UCOS-DA$_{23-\textit{DINOv2}}$\cite{zhang2023unsupervisedco} & .750&.639&.666&.808&.091&.702&.604&.633&.751&.148&.655&.467&..495&.687&.120&.731&.617&.644&.785&.103\\
        \rowcolor{myGray}
        \textbf{Ours}$_{\textit{DINOv1}}$ & .734&.625&.680&.854&.072&.706&.621&.689&.801&.108&.727&.577&.627&.822&.059&.761&.680&.737&.851&.074\\
        \rowcolor{myGray}
        \textbf{Ours}$_{\textit{DINOv2}}$ & \textbf{.864} & \textbf{.825} & \textbf{.838} & \textbf{.931}  & \textbf{.031} & \textbf{.793} & \textbf{.747} & \textbf{.779} & \textbf{.862} & \textbf{.077} & \textbf{.834} & \textbf{.763} & \textbf{.779} & \textbf{.916}  & \textbf{.031} & \textbf{.850} & \textbf{.818} & \textbf{.835} & \textbf{.923} & \textbf{.043} \\
        \bottomrule
	\end{tabular}
 }
 \caption{\textbf{Comparison of our methods with recent methods.} We compared our proposed methods with competing unsupervised, semi-supervised, and full-supervised methods. \textbf{Bold} indicates the best result in group settings, and \underline{underline} indicates the second-best result. \\\textbf{*} denotes the version that reimplemented by us.}\label{tab:main_results}
\end{table*}

\section{Experiment}
\label{sec:exp}

\begin{figure*}[t!]
    \centering
    \includegraphics[width=\linewidth]{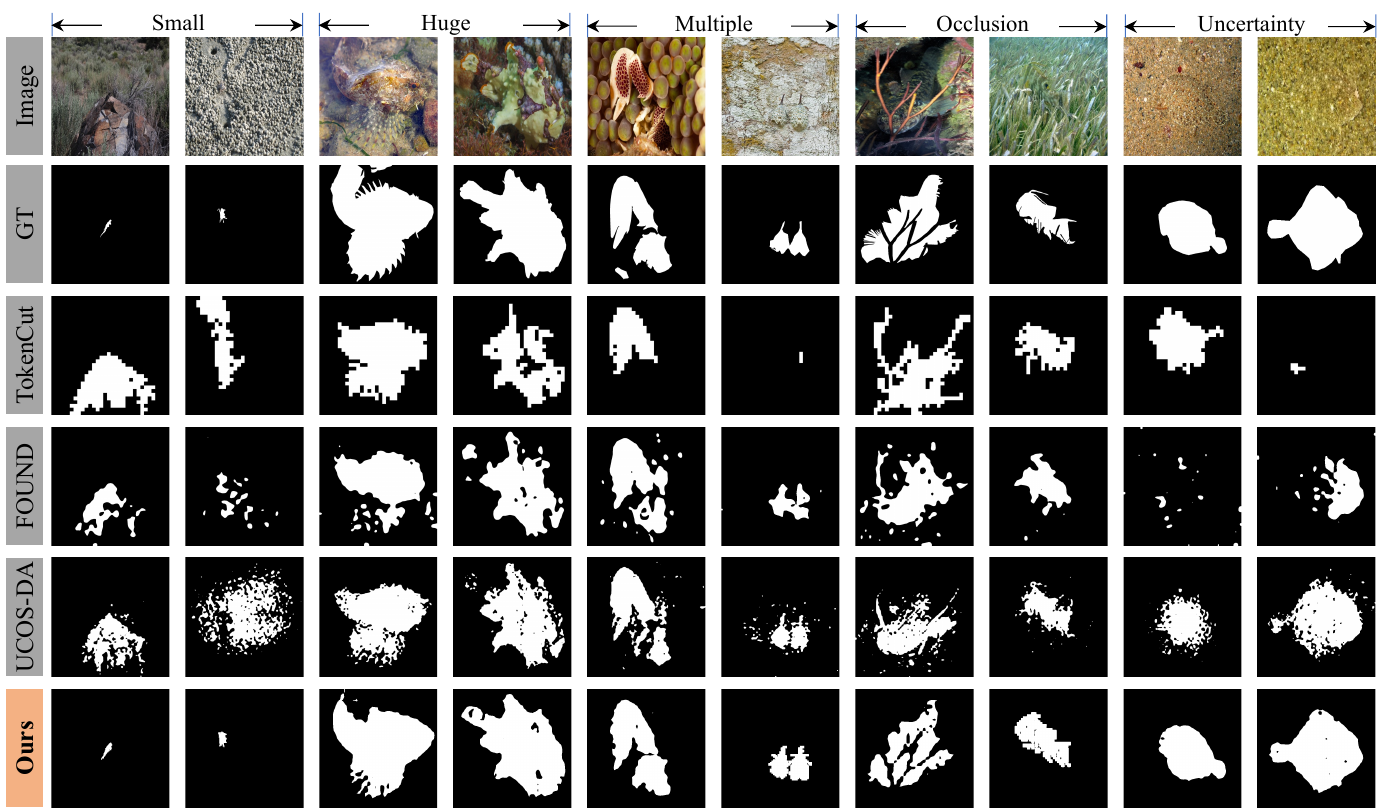}
    \caption{\textbf{Visual comparison of our method with other existing methods in challenging scenarios.} Our method has clearer and more precise segmentation boundaries and correctly recognizes depth-artifacted objects.}
    \label{fig:main_vis}
\end{figure*}

\subsection{Experimental Settings}

\textbf{Trainsets.} To fairly compare with the existing works, following \cite{luo2023camouflaged, fan2020camouflaged}, we used a combination of 1000 images from the CAMO-Train and 3040 images from COD10K-Train as the training dataset of our experiments. Following the unsupervised learning settings, we do not use any ground-truth label during the training process.

\noindent\textbf{Testing Sets.} We evaluate the performance of our methods on four mainstream COD benchmark testing sets: CHAMELEON with 76 test images, CAMO with 250 test images, COD10K with 2026 test images, and NC4K with 4121 test images. 

\noindent\textbf{Evaluation Protocol.} For a fair and comprehensive evaluation, we employ the S-measure ($\mathcal{S}_m$) \cite{fan2017structureanw}, mean and weighted F-measure ($\mathcal{F}_{\beta}^{m}$, $\mathcal{F}_{\beta}^{\omega}$) \cite{margolin2014evaluate}, mean E-measure ($\mathcal{E}_{\epsilon}^{m}$) \cite{Fan2018Enhanced}, mean absolute error ($\mathcal{M}$) \cite{perazzi2012saliency} as the main metrics. 

\noindent\textbf{Implementation Details.} Following \cite{simeoni2023unsupervised}, we employ the simple background seed methods as our fixed strategy branch to generate the coarse pseudo-label. And following \cite{simeoni2023unsupervised,zhang2023unsupervisedco}, we use the strong self-supervised pretrained backbone DINO \cite{caron2021emerging,oquab2023dinov2} as our encoder. The teacher model is updated via EMA with a momentum $\eta$ of 0.99. We train the model for 25 epochs. The batch size is set to 32 for each GPU during training. All experiments are implemented with PyTorch 2.1 and run on a machine with Intel(R) Xeon(R) Silver 4214R CPU @ 2.40GHz, 512GiB RAM, and 1 NVIDIA Titan A100-40G GPUs. All experiments use the same random seed. More implementation details will be provided in the supplementary materials.

\subsection{Main Results}

\noindent\textbf{Qualitative Analysis.}
We show visualizations of a series of camouflaged object segmentation masks predicted by our method and related methods in some challenging scenarios. As shown in \cref{fig:main_vis}, we noticed that our method achieves higher segmentation accuracy than existing approaches, with clearer edges and a more detailed representation of the camouflaged object. Note that \textit{DINOv2} is employed for all methods that require a backbone to extract features.

\noindent\textbf{Quantitative Analysis.}
In \cref{tab:main_results}, we compared our proposed method's performance with competing USS and UCOD models on four COD test datasets. To further highlight our model's competitiveness, we also compared existing fully-supervised and semi-supervised methods. The results show that our model outperformed all existing USS and UCOD methods across all metrics and datasets, thus achieving State-Of-The-Art (SOTA) performance. Additionally, based on \textit{DINOv2} backbone, our model has surpassed several semi-supervised and fully-supervised methods across all four datasets, demonstrating its superior performance, effectiveness, and robustness. 

\subsection{Ablation Study}
To verify the effectiveness of our methods, we employ \textit{DINOv2} as the backbone and conduct comprehensive ablation studies on the COD10K-Test dataset with 2,026 images. 

\begin{table}[]
    \setlength{\belowcaptionskip}{0cm}   
    \renewcommand{\arraystretch}{1.1}
    \renewcommand{\tabcolsep}{3pt}
    \centering
    \resizebox{\linewidth}{!}{
    \begin{tabular}{cccc|ccccc}
    \toprule
    \multicolumn{4}{c|}{\textbf{Settings}}  & \multicolumn{5}{c}{\textbf{COD10K (2026)}} \\ \midrule
    \textbf{Tea-Stu} & \textbf{APM} & \textbf{DBA} & \textbf{Look-Twice} \metricsCOD{}  \\ \midrule
    \checkmark &            &            &            &  .626 & .448 & .464 & .658 & .112 \\
    \checkmark &            &            & \checkmark &  .633 & .452 & .472 & .668 & .129 \\
               &            & \checkmark &            &  .607 & .409 & .435 & .631 & .129 \\
               &            & \checkmark & \checkmark &  .603 & .404 & .431 & .625 & .133 \\
    \checkmark & \checkmark &            &            &  .724 & .582 & .604 & .800 & .054 \\
    \checkmark & \checkmark &            & \checkmark &  .788 & .679 & .700 & .879 & .043 \\
    \checkmark & \checkmark & \checkmark &            &  .784 & .675 & .692 & .869 & .036 \\ \rowcolor{myGray}
    \checkmark & \checkmark & \checkmark & \checkmark & \textbf{.834} & \textbf{.763} & \textbf{.779} & \textbf{.916}  & \textbf{.031} \\
     \bottomrule
    \end{tabular}}
    \caption{\textbf{Ablation study to evaluate the proposed modules.} We retrained our model with different settings on the same learning rate and epochs, then benchmarked on COD10K-testset to validate the effectiveness.}
    \label{tab:abl_main}
\end{table}

\begin{table}[]
    \setlength{\belowcaptionskip}{0cm}   
    \renewcommand{\arraystretch}{1.1}
    \renewcommand{\tabcolsep}{7pt}
    \centering
    \resizebox{\linewidth}{!}{
    \begin{tabular}{c|cccccc}
    \toprule
    \multirow{2}*{\textbf{Mixing Strategies}} & \multicolumn{5}{c}{\textbf{COD10K (2026)}} \\ \cmidrule[0.05em](lr){2-6}
     \metricsCOD{}  \\ \hline
     Proportional(1:1) Mixing & .707& .555& .572& .763& .082  \\
     Linear Decay Mixing      & .809& .716& .723& .888&.038  \\\rowcolor{myGray}
     APM                      & \textbf{.834} & \textbf{.763} & \textbf{.779} & \textbf{.916}  & \textbf{.031}       \\  
     \bottomrule
    \end{tabular}}
    \caption{\textbf{Ablation to the pseudo-label mixing strategy.} We retrained our model with different pseudo-label mixing strategies and benchmarked on the COD10K-Test dataset.}
    \label{tab:abl_mixing}
\end{table}

\begin{figure}
    \centering
    \includegraphics[width=\linewidth]{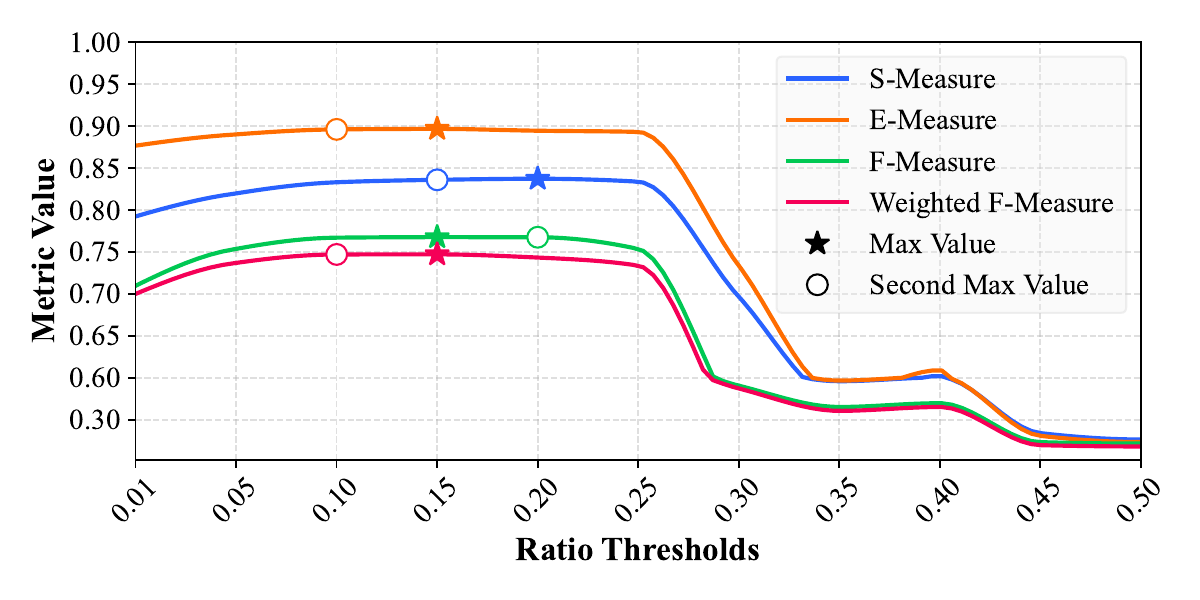}
    \caption{\textbf{Hyper-parameter ablation of small-sized object ratio $\tau$.} The model performs optimally and shows stable results when the value of $\tau$ is set around 0.15.}
    \label{fig:tau_abl}
\end{figure}

\begin{table}[]
    \setlength{\belowcaptionskip}{0cm}   
    \renewcommand{\arraystretch}{1.1}
    \renewcommand{\tabcolsep}{6pt}
    \centering
    \resizebox{\linewidth}{!}{
    \begin{tabular}{c|cccccc}
    \toprule
    \multirow{2}*{\textbf{Fixed Generation Strategies}} & \multicolumn{5}{c}{\textbf{COD10K (2026)}} \\ \cmidrule[0.05em](lr){2-6}
     \metricsCOD{}  \\ \hline
     Null(Pure Black/White) &  .459 & .002 & .001 & .250 & .091 \\
     Random Perlin Noise &  .667 & .489 & .543 & .788 & .072\\
     \rowcolor{myGray}
     Background Seed         & \textbf{.834} & \textbf{.763} & \textbf{.779} & \textbf{.916}  & \textbf{.031}       \\  
     \bottomrule
    \end{tabular}}
    \caption{\textbf{Fixed Pseudo-label Generation Strategy Ablation.} We retrained our model with a replaced fixed strategy and benchmarked it on the COD10K-Test dataset.}
    \label{tab:abl_plabel}
\end{table}

\noindent\textbf{Module Ablation.} We first conduct ablation experiments on the teacher-student framework (Tea-Stu) and our proposed methods: Adaptive Pseudo-label Mixing (APM) module, Dual-Branch Adversarial (DBA) decoder, and Look-Twice strategy. To thoroughly demonstrate their effectiveness, we replaced the decoder with a single $1 \times 1$ convolutional layer following \cite{simeoni2023unsupervised, zhang2023unsupervisedco}. For all ablations on the APM module, we used the $1:1$ constant mixing strategy to replace it. The ablation results are shown in \cref{tab:abl_main}. 
As a result, we find that when only the teacher-student framework is used, the simple $1 \times 1$ convolution fails to learn camouflage-related knowledge adequately. When training with a DBA decoder only, the model lacks the guidance of the teacher-student framework, thus easily learning the incorrect knowledge from the low-quality pseudo-labels constructed by fixed-strategy. The proposed Look-Twice mechanism is effective only when the network can produce sufficiently high-quality predictions. If the network has poor localization and segmentation capabilities, Look-Twice will degrade performance.

\noindent\textbf{Mixing Strategy Ablation.} We continue conducting ablation studies on the mixing strategies. We compared the proportional(1:1), linear decay, and APM mixing strategies. As shown in \cref{tab:abl_mixing}, the APM module provides a method for fusing pseudo-labels that is more aligned with the model's training process and achieving optimal performance.

\noindent\textbf{Hyper-parameter Ablation.} We conducted ablation study on the threshold $\tau$ for the small-sized object ratio mentioned in \cref{subsec:look_twice}. As shown in \cref{fig:tau_abl}, when $\tau \approx 0.15$, the model achieves optimal and stable performance. Therefore, we set $\tau$ to 0.15 in the final model.

\begin{figure}[t!]
    \centering
    \includegraphics[width=\linewidth]{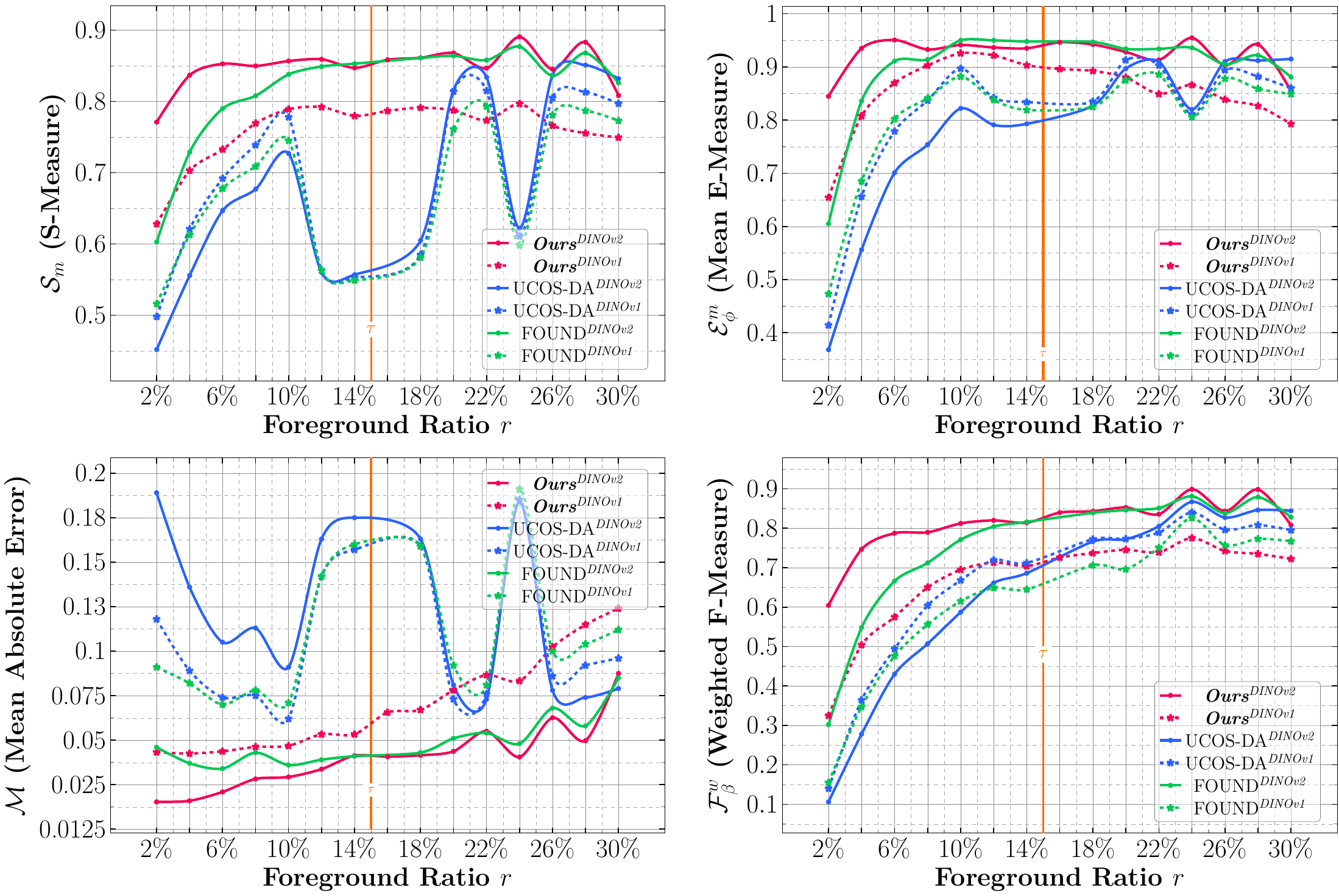}
    \caption{\textbf{Performance comparison for different foreground sizes on COD10K-Test dataset.}}
    \label{fig:fore_abl}
\end{figure}

\noindent\textbf{Study on Foreground-sizes.} We divided the test set by the proportion of test set prospects at 2\% intervals, then benchmarked the performance between our method and some previous SOTA methods\cite{simeoni2023unsupervised, zhang2023unsupervisedco}. As shown in \cref{fig:fore_abl}, our method has significant advantages when dealing with smaller-sized camouflaged targets.

\noindent\textbf{Fixed Pseudo-label Generation Strategy Ablation.} We further investigated the effect of the pseudo-labels constructed by the fixed-strategy. For each input image, we generated a random shape of Perlin noise on a plain black background. We used this ``fabricated pseudo-label" as well as the pure black and white masks to replace the fixed-strategy pseudo-label during the training process. The results are shown in \cref{tab:abl_plabel}. When using random noise, the noise may overlap with the foreground region, allowing the network to explore some knowledge related to the camouflaged foreground. However, when using a solid color mask, the model fails to learn the basic definition of the segmentation task to produce valid predictions.

\section{Conclusions}
\label{sec:conclusions}

Considering the existing UCOD methods suffer from main issues: 1). The low-quality pseudo-labels constructed by fixed strategies contain substantial noise, making the model prone to fitting incorrect knowledge. 2). The low-resolution pseudo-labels cause severe confusion between the foreground and background. Simple $1 \times 1$ convolution fails to capture and learn the semantic features of camouflaged objects, especially for small-sized samples. To solve these problems, we propose UCOD-DPL in this paper. We introduce a teacher-student framework with an Adaptive Pseudo-label Module, a Dual-Branch Adversarial decoder, and a Look-Twice mechanism. 
The APM dynamically fuses pseudo-labels from the fixed strategy and teacher model, effectively preventing the model from overfitting the noise and erroneous knowledge. The DBA decoder mines foreground and background knowledge relevant to camouflaged objects while resisting incorrect information from pseudo-labels. The Look-Twice refinement mechanism enhances camouflaged object learning and refinement of small objects.
We conducted extensive benchmarking on multiple COD test datasets, and the results show that our method outperforms all previous approaches and even reaches the performance level of some fully supervised methods.

\section*{Acknowledgements}
This work was supported by the National Science Fund for Distinguished Young Scholars (No.62025603), the National Natural Science Foundation of China (No. U21B2037, No. U22B2051, No. U23A20383, No. 62176222, No. 62176223, No. 62176226, No. 62072386, No. 62072387, No. 62072389, No. 62002305 and No. 62272401), and the Natural Science Foundation of Fujian Province of China (No. 2021J06003, No. 2022J06001).

{
    \small
    \bibliographystyle{ieeenat_fullname}
    \bibliography{main}
}


\end{document}